\documentclass[lettersize,journal]{IEEEtran}
\usepackage{algorithmic}
\usepackage{algorithm}
\usepackage{array}
\usepackage{textcomp}
\usepackage{stfloats}
\usepackage{url}
\usepackage{verbatim}
\usepackage{graphicx}
\usepackage{cite}
\usepackage{amsmath,amssymb,amsfonts}
\usepackage{xcolor}
\usepackage{subfigure}
\usepackage{multirow}
\usepackage{booktabs}
\usepackage{hyperref}
\usepackage{authblk}
\usepackage{colortbl}
\graphicspath{{fig/}} %figure path

\def\tablename{Table}
\newcommand{\CLA}[1]{{\color[HTML]{B84232} \textbf{#1}}}
\newcommand{\CLB}[1]{{\color[HTML]{2A4A8C} \textbf{#1}}}

\begin{document}

\title{Embodied Image Quality Assessment for Robotic Intelligence}

\author{Jianbo~Zhang$^{1}$, Chunyi~Li$^{1}$, Jie~Hao$^{2}$, Jun~Jia$^{1}$, Huiyu~Duan$^{1}$, Guoquan~Zheng$^{2}$, Liang~Yuan$^{3}$, Guangtao~Zhai$^{1}$,~\IEEEmembership{Fellow,~IEEE}
        % <-this % stops a space
\thanks{This work was supported by the National Key Research and Development Program of China under Grant 2023YFB4704000, and in part by the National Natural Science Foundation of China under Grant 52275003. ($Corresponding \  author: \  Liang \  Yuan, Guangtao~Zhai$.)}% <-this % stops a space
\thanks{Jianbo Zhang, Chunyi Li, Jun Jia, Huiyu Duan, and Guangtao~Zhai are with the Institute of Image Communication and Network Engineering, Shanghai Jiao Tong University, Shanghai 200240, China (E-mail: \{sjtu5029101, lcysyzxdxc, jiajun0302, huiyuduan, zhaiguangtao\}@sjtu.edu.cn)}
\thanks{Jie Hao and Guoquan Zheng are with the College of Information Science and Technology, Beijing University of Chemical Technology, Beijing 100029, China (E-mail: haojie@buct.edu.cn, quanmoxiansheng@163.com)}
\thanks{Liang Yuan is with USC-SJTU Institute of Cultural and Creative Industry, Shanghai Jiao Tong University, Shanghai 200240, China (E-mail: lyuan@sjtu.edu.cn)}}

% \markboth{Journal of \LaTeX\ Class Files,~Vol.~ , No.~ ,  ~20 }%

% \IEEEpubid{0000--0000/00\$00.00~\copyright~20xx IEEE}

\maketitle

\begin{abstract}
Image Quality Assessment (IQA) of User-Generated Content (UGC) is a critical technique for human Quality of Experience (QoE). However, does the the image quality of Robot-Generated Content (RGC) demonstrate traits consistent with the Moravec paradox, potentially conflicting with human perceptual norms? 
Human subjective scoring is more based on the attractiveness of the image. Embodied agent are required to interact and perceive in the environment, and finally perform specific tasks. Visual images as inputs directly influence downstream tasks.
In this paper, we explore the perception mechanism of embodied robots for image quality. We propose the first Embodied Preference Database (EPD), which contains 12,500 distorted image annotations. We establish assessment metrics based on the downstream tasks of robot. In addition, there is a gap between UGC and RGC. To address this, we propose a novel Multi-scale Attention Embodied Image Quality Assessment called MA-EIQA. For the proposed EPD dataset, this is the first no-reference IQA model designed for embodied robot.
Finally, the performance of mainstream IQA algorithms on EPD dataset is verified. The experiments demonstrate that quality assessment of embodied images is different from that of humans.  
We sincerely hope that the EPD can contribute to the development of embodied AI by focusing on image quality assessment. The benchmark is available at \url{https://github.com/Jianbo-maker/EPD_benchmark}. 
\end{abstract}


\section{Introduction}
\IEEEPARstart{O}{ver} the past two decades, Image Quality Assessment (IQA) has undergone significant evolution. Traditional methods are designed based on the Human Visual System (HVS) and focus on human subjective experience of images \cite{IQASURVEY,TCSVT-IQA1,TCSVT-IQA2}. However, how does an Embodied Artificial Intelligence (EAI) assess image quality based on their own experience? The EAI emphasizes interaction and perception in the real environment. Robots perform specific tasks through interaction with their surroundings, where image quality is crucial in determining the success of these tasks. In addition, the robotics community is focused on the development of denoising algorithms for different types of image noise, including dehazing, deraining, and deblurring \cite{dehazing,deraining,debluring}. The assumption based on the same type of image distortion is not flexible enough, and it also lacks exploration of the relevance of image quality to embodied tasks. The development of a robot-centric image quality assessment is important for the technology of visual signal processing and embodied robots.

\begin{figure}[htbp]
  \centering
  \includegraphics[width=8cm]{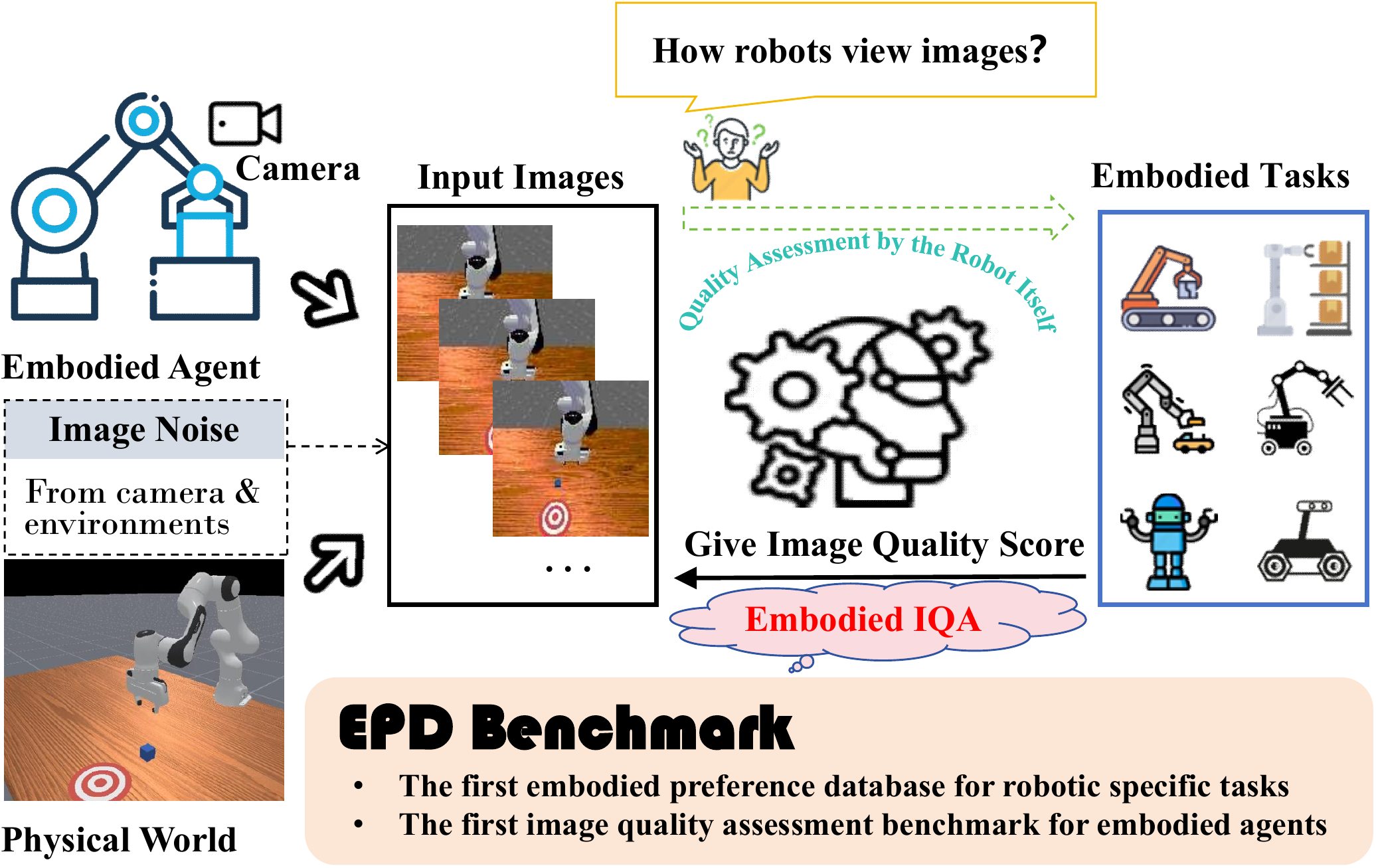}
  \caption{Input image quality is assessed based on the performance of the robot for embodied tasks. The EPD benchmark benefit image quality assessment for EAI.}
  \label{abstract}
\end{figure}

Unlike offline machine learning, which is driven by massive amounts of internet data, embodied AI currently faces a severe bottleneck due to a lack of high-quality data. Consequently, by understanding how image quality affects downstream tasks, an embodied robot can autonomously determine which image data are beneficial for its task during acquisition and usage, and thus improving its overall performance.

There are inherent disparities between the HVS and the Robot Visual System (RVS). Generally, humans prioritize the high-level semantic content of an image, whereas machines tend to analyze the low-level geometric structure. Robots prioritize the consistency of image texture and structure and are less sensitive to changes in semantic information, which contrasts with human perceptions of image quality.

Traditional IQA methods are primarily designed based on the HVS \cite{TCSVT-IQA3,TCSVT-IQA4}. These model architectures are relatively large, which are difficult to apply to embodied AI agents. Embodied robot operate under strict computational resource constraints and require real-time responses. Furthermore, the visual perception tasks of embodied AI are more focused on accurately understanding physical information within a scene, such as texture details, edge outlines, and structural integrity, rather than the aesthetic factors central to human perception.

Vision-based embodied artificial intelligence has a wide range of research prospects. However, a comprehensive framework for assessing the image quality of robotic vision systems remains undeveloped. The application of vision technology to robotics itself has inherent limitations. Vision cameras are passive sensors that passively receive information from the environment. The camera itself and its surroundings are subject to a large number of interfering factors, especially in the image collection and transmission where additional noise is introduced. Embodied methods designed only through desirable image assumptions are difficult to interact effectively with the real world. The primary bottleneck that hinders this research is the absence of image quality assessment for embodied AI. Considering this, this paper proposes a benchmark and an approach for quality assessment of embodied images. Fig. \ref{abstract} illustrates the paradigm of the proposed benchmark. In addition, given the inherent differences between HVS and RVS, traditional IQA approaches are not applicable to the new scenario. Therefore, we propose an embodied image quality assessment method for this new area. Our main contributions are as follows:

\begin{figure*}[t] %htbp
  \centering
  \includegraphics[width=18cm]{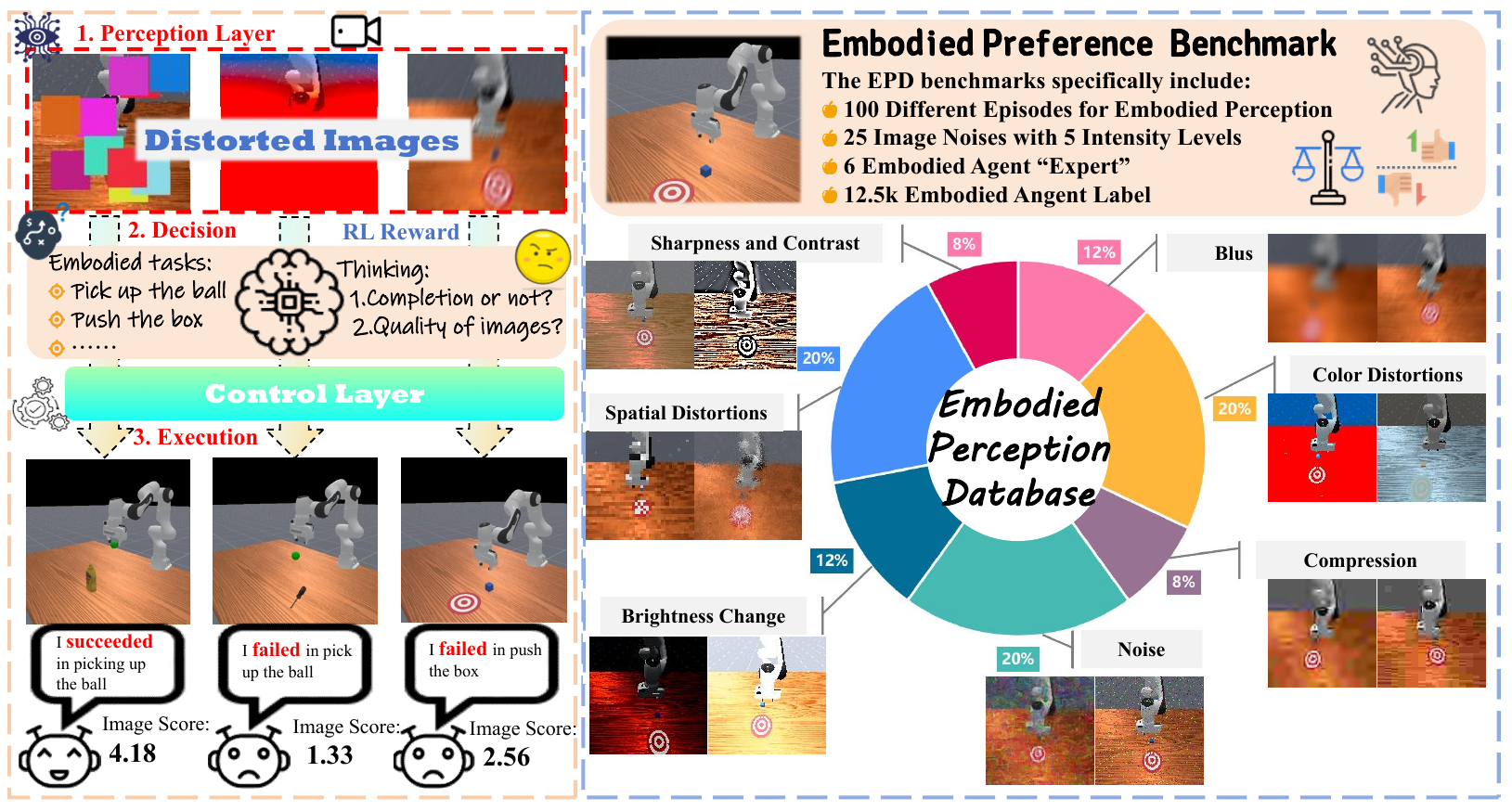}
  \caption{The construction overview of the proposed EPD Benchmark. The EPD benchmark is oriented towards embodied AI and includes 25 common image distortions with 5 different levels of intensity. There are 100 pairs of reference/distortion images for each distortion type.}
  \label{flowchart}
\end{figure*}

\definecolor{level-1}{HTML}{66B366}
\definecolor{level-2}{HTML}{E2B34B}
\definecolor{level-3}{HTML}{9A70C2}
\definecolor{level-4}{HTML}{7AADC8}
\definecolor{level-5}{HTML}{C87A7A}

\begin{figure*}[t] %htbp
  \centering
  \includegraphics[width=16cm, height=12.8cm]{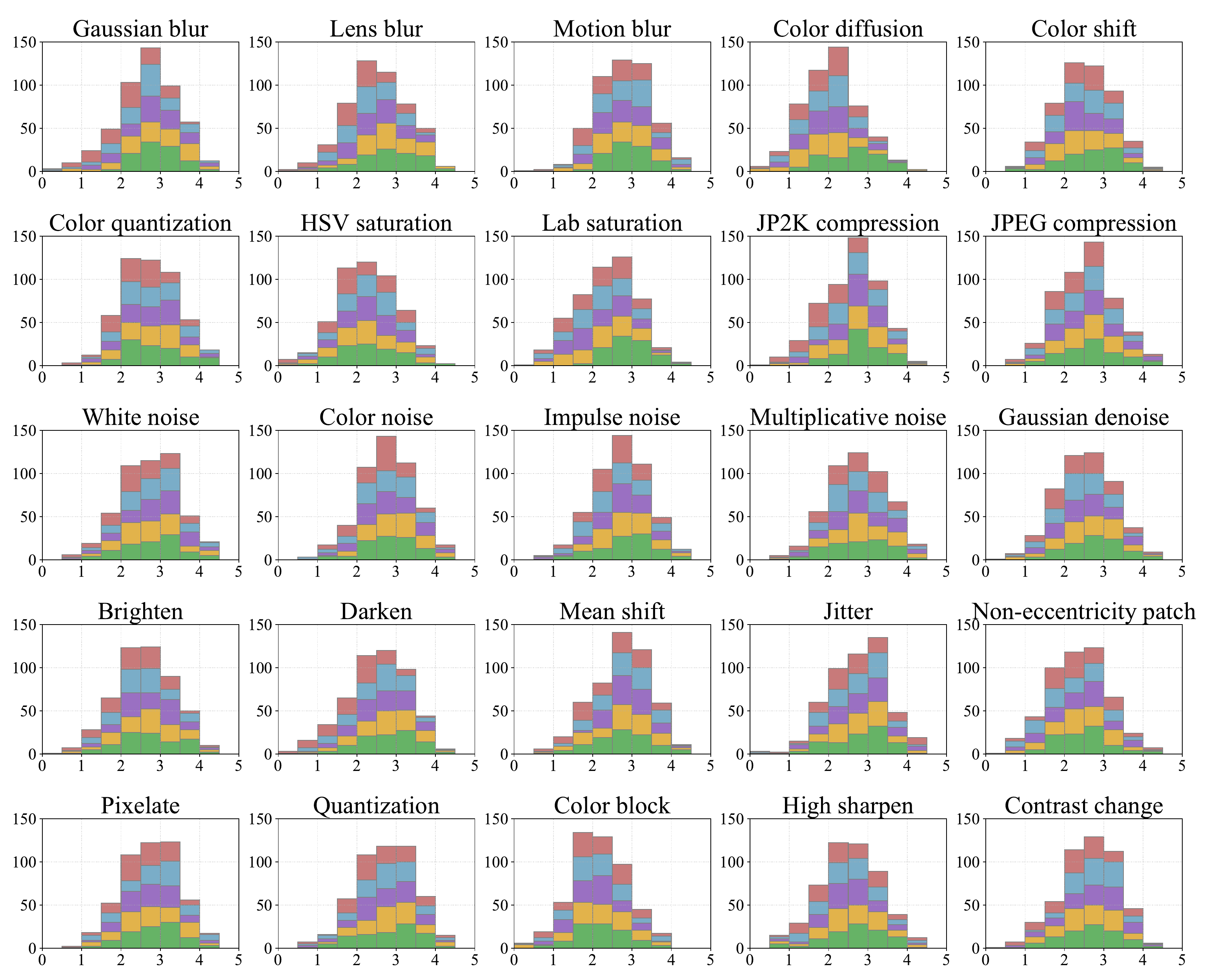}
  \caption{DMOS of the EPD, visualized in 25 distortion subsets. Different color denotes distortion strength \textcolor{level-1}{Level 1} , \textcolor{level-2}{Level 2}, \textcolor{level-3}{Level 3}, \textcolor{level-4}{Level 4}, \textcolor{level-5}{Level 5}. Distribution results show the sensitivity of the embodied robot to each distortion varied significantly.}
  \label{DMOS}
\end{figure*}

\begin{itemize}
   \item We construct the embodied preference database EPD, which is the first IQA database for embodied AI. This database comprises 12,500 reference/distorted image pairs. The images are annotated via a robotic arm completing the embodied tasks without any human participation. The EPD explore the internal mechanism of IQA in embodied AI.
   \item We propose the first embodied image quality assessment model, which is Multi-scale Attention Embodied Image Quality Assessment (MA-EIQA) network. MA-EIQA exceeds the performance of traditional complex IQA networks only through its lightweight structure.
   \item We benchmarked existing IQA methods in the EPD database. The experimental results validated the existence of inconsistency between HVS and RVS. 
 \end{itemize}

% \subsection{Lists}
% In this section, we will consider three types of lists: simple unnumbered, numbered, and bulleted. There have been many options added to IEEEtran to enhance the creation of lists. If your lists are more complex than those shown below, please refer to the original ``IEEEtran\_HOWTO.pdf'' for additional options.\\

% \begin{list}{}{}
% \item{bare\_jrnl.tex}
% \item{bare\_conf.tex}
% \item{bare\_jrnl\_compsoc.tex}
% \item{bare\_conf\_compsoc.tex}
% \item{bare\_jrnl\_comsoc.tex}
% \end{list}

\section{Related Work}
\subsection{Image Quality Assessment}
Human-oriented image quality assessment methods can be divided into three categories, the Full-Reference (FR) IQA, the Reduced-Reference (RR) IQA, and the No-Reference (NR) IQA. 

The FR IQA method assumes that a distortion-free reference image exists and that the reference image information is fully available. Early approaches are based primarily on image content and and structure at the pixel level. SSIM~\cite{ssim} combines contrast information, image brightness information, and image structural similarity information to estimate the quality of an image. A number of subsequent work has built on this foundation to continue to advance IQA, such as VW-SSIM~\cite{VW_SSIM}, IW-SSIM~\cite{IW_SSIM}, VIF~\cite{VW_SSIM}, FSIM~\cite{FSIM}, GMSD~\cite{GMSD}. In recent years, the FR-IQA approach incorporating deep learning has yielded more accurate evaluation results~\cite{FR-IQA1,FR-IQA2}. 

The RR IQA method usually assesses the quality of an image using partial information from a reference image or extracting a small number of features~\cite{RR-IQA}. The NR IQA approach usually refers to the assessment of distorted image quality in the absence of the original distortion-free image~\cite{NR-IQA1,NR-IQA2}. Methods based on Natural Scene Statistics (NSS) features have been studied in the early stages. Deep learning-based algorithms for reference-free quality evaluation are gradually replacing traditional methods~\cite{DLIQA1,DLIQA2}.

In addition, the benchmark for image quality assessment is a crucial issue. Synthetic distortion datasets such as LIVE ~\cite{LIVE} and KADID-10k  ~\cite{KADID-10k} are used to evaluate traditional image quality evaluation algorithms. In order to eliminate the gap between synthetic distorted images and distorted images in the wild, datasets such as LIVEC ~\cite{LIVEC} and KonIQ-10k ~\cite{KonIQ-10k} have been constructed. In recent years, IQA evaluation benchmarks for multi-modal large language models (MLLMs) are proposed. Q-Bench++ ~\cite{Q-Bench++} evaluated the performance of MLLMs on images in three aspects of perception, description, and assessment. AI-Generated Image (AGI) subjective quality assessment database and benchmark are proposed ~\cite{AGIQA-3K,A-Bench,CMC-Bench}. R-bench ~\cite{R-Bench} evaluates the robustness of MLLMs in the real-world. Previous work has modeled the Human Visual System (HVS) for image quality assessment. However, there are differences in how humans and embodied robots view images. To the best of our knowledge we present the first image quality assessment for Embodied Artificial Intelligence. 

\subsection{Embodied Artificial Intelligence}

In contrast to traditional deep learning, embodied artificial intelligence emphasizes the exploration of the surrounding environment by an intelligent agent, which is subject to active perception, interaction, and reasoning. Existing work studies embodied intelligence from a number of different perspectives, such as embodied agent, embodied simulators, embodied perception, embodied interaction~\cite{Embodied-AI-survey}.

With the aim of being able to actively interact with the real world beyond mere simulation in a virtual world on a computer, different robotic bodies help the AI to have a physical body that is concrete. Wheeled and tracked robots are characterized by simplicity of structure and mobility, but are limited in complex terrain~\cite{wheel_robots1,wheel_robots2}. The biomimetic robots achieve movement in different scenarios by simulating the movement characteristics of living creatures~\cite{bio_robot1,bio_robot2,bio_robot3,bio_robot4}. The mechanical structure and motion control of these robots are more complicated, and they are less reliable and stable in real scenarios. As the most mature product in industrial scenarios, fixed-base robots have gained a great deal of research due to their stability and high operational precision~\cite{base_robot1,base_robot2}. This paper explores the extent to which image quality affects embodied tasks using a fixed-base robotic arm task for embodied robots, with more explorations of embodied intelligences to be completed subsequently.

As with the development of traditional AI, embodied artificial intelligence also requires a large amount of experimental data and scenarios for testing. In general, a simulator that can simulate the real physical world can provide an efficient and scalable experimental environment, subject to the limitations of experimental conditions and considerations of experimental safety. The Issac Sim~\cite{Isaac}, Gazebo~\cite{Gazebo}, and PyBullet~\cite{pybullet} provide a general-purpose simulator to support algorithm development and model training. Simulators such as AI2-THOR~\cite{AI2-THOR}, Habitat~\cite{Habitat}, SAPIEN~\cite{SAPIEN}, etc. provide a large number of scenarios that simulate real-world, mission-specific scenarios, including virtual scenes and virtual objects. In this paper, the SAPIEN is used as a simulation environment to evaluate the image quality by using a robotic arm to perform specific tasks.

\section{Database Construction}

\subsection{Image Collection}

In contrast to traditional IQA image collection methods, embodied AI requires interaction with the surrounding environment. The ultimate goal is robot-oriented image quality assessment, and thus, image collection is also done by the robot itself. Two classical reinforcement learning algorithms, the Proximal Policy Optimization (PPO) \cite{PPO} and the Soft Actor-Critic (SAC) \cite{SAC}, and a state-of-the-art method, TDMPC2 \cite{TDMPC2}, are used to perform the 2 tasks in the SAPIEN simulator \cite{SAPIEN}, respectively. A monocular camera is used to capture RGB images as sensor data input to the model.

\begin{figure}[t] %hbp
  \centering
  \subfigure[Correlation matrix]{
    \begin{minipage}{4cm}
    \centering
        \includegraphics[width=4cm,height=3.5cm]{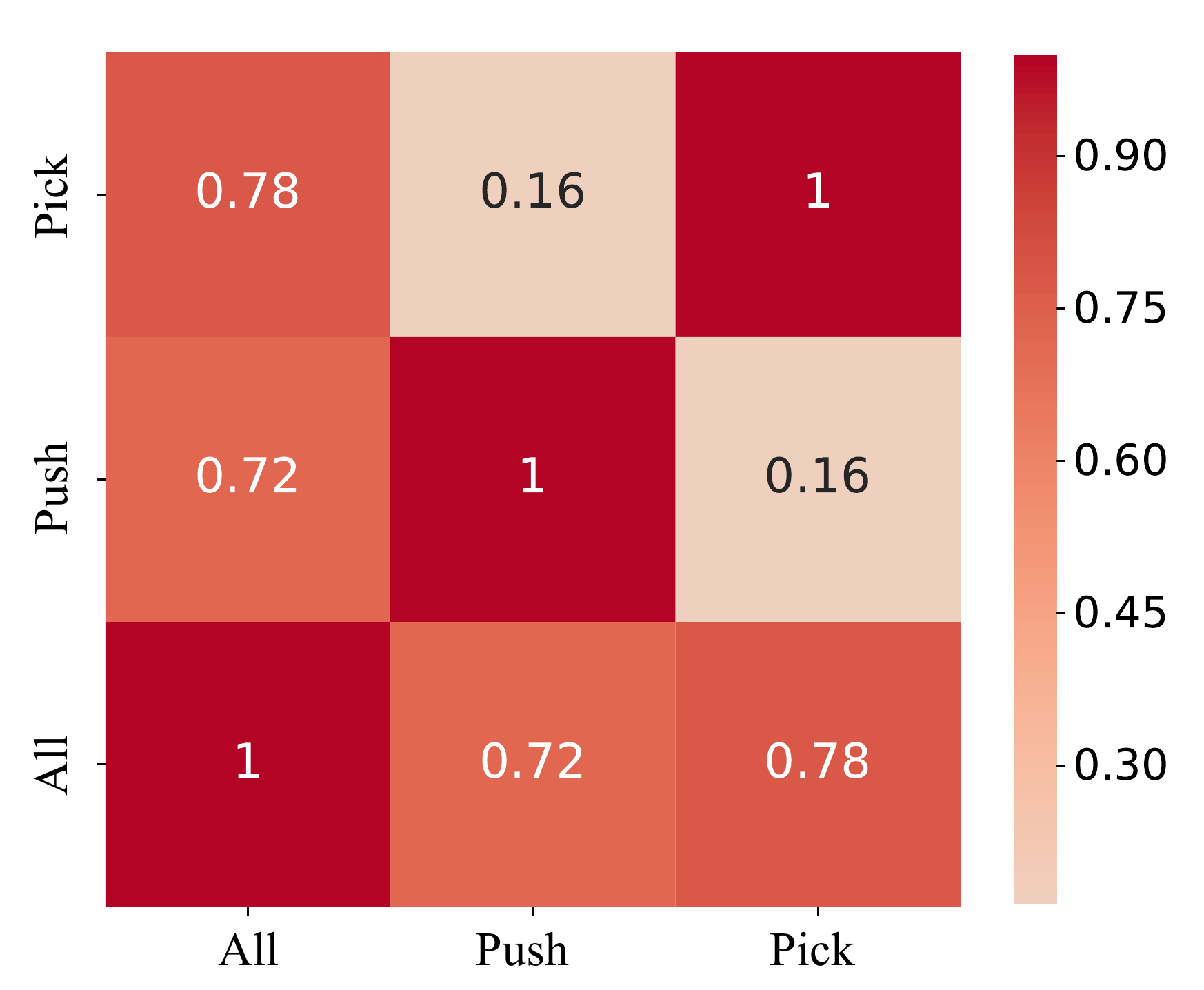}
    \end{minipage}}
    \subfigure[Preference score]{
      \begin{minipage}{4cm}
      \centering
          \includegraphics[width=4cm,height=3.5cm]{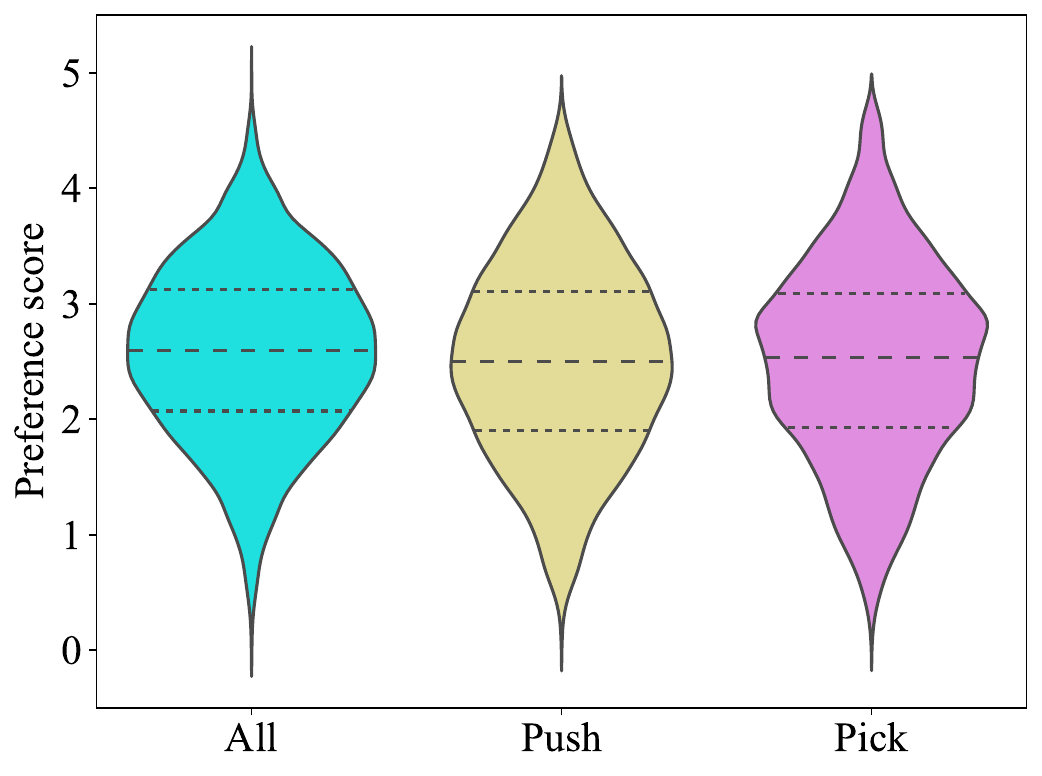}
      \end{minipage}}
  \caption{The dataset analyses of EPD. (a) The correlation matrix for MOS. It is the correlation analyses for the subsets of the all task, the push task and the pick task respectively. (b) The distribution of nomalized data. It can be seen that the data distribution is similar among different tasks.}
  \label{data relation}
\end{figure}

For distorted images, 25 common types of image distortions with 5 different levels of intensity are set by modelling the image inputs by simulating camera distortions and environmental distortions. Specifically, these image distortion types are classified into 7 categories are Blur distortions, Color distortions, Compression, Noise, Brightness change, Spatial distortions, Sharpness and contrast. A embodied task requires multiple images to work together, but the image distortion type is set to be consistent. Therefore, only the image of the initial frame is selected as the image to be evaluated. For a more extensive evaluation, 100 different initial scenes are set for each task separately. In summary, a total of 12,500 reference and distorted images are collected for the database. The construction overview of the proposed EPD benchmark is shown in Fig. \ref{flowchart}.

\subsection{Preference Score Collection}
Based on a simulated environment, a robotic arm acts as an embodied intelligence to perform simple push and pick tasks. For the robot, different quality of image inputs have different impacts on the robot to complete the task, which is also directly related to the performance of the robot. Based on the ManiSkill platform \cite{ManiSkill3}, three reinforcement learning algorithms are used to complete two different tasks in the same scene. The reward value of each episodes is adopted as the performance score for evaluating the robot performance in the process of perceiving the environment and completing the embodied intelligence task. 
In addition, the evaluation of RL tasks can be based on reward values, success rates, and accuracy. However, success rates and accuracy cannot provide as detailed an assessment of task performance as reward values, which record scores for each step of a task.
The robotic arm perceives the surrounding environment, then makes decisions based on the perception results, and finally executes the corresponding actions.

For the PPO algorithm \cite{PPO}, the goal of the robot is to find a policy parameter $\theta$ that maximises the expected cumulative reward. The reward function $r(\mathbf{s}_{t}, \mathbf{a}_{t})$ represents the immediate reward obtained by the agent performing action $\mathbf{a}_{t}$ in state $\mathbf{s}_{t}$ at time step $t$. PPO maximises the cumulative reward in the environment by optimising the policy $\pi_{\theta}$. The expectation of the cumulative reward $J(\theta)$ is denoted as:
\begin{equation}
  J(\theta)=E_{\tau\sim \pi_\theta(\tau)}
  \left[\sum_{t=1}^{T} r(\mathbf{s}_t,\mathbf{a}_t)\right],
\end{equation}
where the $E_{\tau\sim \pi_\theta(\tau)}[\cdot]$ denotes the expected value of the cumulative reward $r(\tau)$ for all possible trajectories $\tau$ (i.e., sequences of states and actions). In addition, the trajectory $\tau$ is generated by the strategy $\pi_{\theta}$.

The SAC \cite{SAC} method introduces an entropy regularisation term in reinforcement learning to encourage the agent to explore more actions. The reward function of SAC contains not only the immediate reward of the environmental feedback but also the entropy value of the current strategy in order to balance the exploration with the expected reward. Denote immediate reward from $\mathrm{s}_t$ to $\mathrm{s}_{t+1}$ as $\mathrm{R}(\mathrm{s}_t,\mathrm{a}_t,\mathrm{s}_{t+1})$. Given the $\mathrm{H}(\pi(\cdot\mid\mathrm{s}_t))$ denotes the entropy of the strategy $\pi$ in state $s_t$. We denote the expectation reward function of SAC as $\pi^*$:
\begin{equation}
  \begin{aligned}
    \pi^*= &\arg\max_\pi E_{\tau\sim\pi}\\
  & \left[\sum_{\mathrm{t}=0}^\infty\gamma^\mathrm{t}\left(\mathrm{R}\left(\mathrm{s}_t,\mathrm{a}_t,\mathrm{s}_{t+1}\right)+\alpha\mathrm{H}\left(\pi\left(\cdot\mid\mathrm{s}_t\right)\right)\right)\right],
  \end{aligned}
\end{equation}
where $\gamma^\mathrm{t}$ is scale factor, and $\alpha$ is entropy regularisation factor. 

TDMPC2 \cite{TDMPC2} derives its closed-loop control strategy by planning with the learnt world model. The agent utilises the model to predict future states and selects the optimal sequence of actions based on the predictions. We denote $\mathbf{D}(\mathbf{s}_{t+1},\mathbf{s}_\mathrm{goal~})$ as the distance between $\mathbf{s}_{t+1}$ and $\mathbf{s}_\mathrm{goal~}$. Therefore, the reward function of TDMPC2 $\mathbf{W}(\mathbf{s}_t,\mathbf{a}_t,\mathbf{s}_{t+1})$ is:
\begin{equation}
  \mathbf{W}(\mathbf{s}_t,\mathbf{a}_t,\mathbf{s}_{t+1})=r(\mathbf{s}_t,\mathbf{a}_t)+\lambda\cdot\mathrm{D}(\mathbf{s}_{t+1},\mathbf{s}_\mathrm{goal~})^{-1},
\end{equation}
where $\lambda$ is the weight factor.

Each episodes receives a reward corresponding to the score of the image, and finally the embodied Differential Mean Opinion Score (DMOS) are normalised to the range (0, 5).

\subsection{Database Analysis}
Fig. \ref{DMOS} illustrates the DMOS distribution of the 25 distortions with 5 levels. From the results, it can be seen that the assessment of an image by a robot is not the same as the assessment of an image by a human. Robots are not as sensitive as humans to changes in different levels of intensity when performing tasks. Essentially, embodied robots focus on the low-level pixel-level features of an image, while humans focus on the high-level semantic features of an image. Therefore, the type of distortion that affects the pixel features can seriously affect the embodied task. Whereas the type of distortion that destroys the semantic integrity, which is more of a concern to the human visual system, only affects the image assessment of humans. Hence the method of image quality assessment that based on human visual system is not work well in EPD.

The DMOS of the EPD and the subscores of the two tasks are shown in Fig. \ref{data relation}, and their correlation is quantified by the Spearman Rank Order Correlation Coefficient (SRCC). As can be seen from the figure, the correlation between tasks is low. Since the difficulty of both push and pick tasks is different, the reward values for the tasks are also different. The data distributions for the two tasks and the total DMOS are similar, both showing an approximate normal distribution. In addition, there is a strong correlation between the overall scores and the individual tasks, with all SRCC exceeding 0.5. The EPD dataset is reliable in this category of tasks.

% \begin{figure}[t]
%   \centering
%   \includegraphics[width=8cm]{distort_distri.pdf}
%   \caption{MOS score of the EPD, visualized in 25 distortion subsets. The results show the reflection of different image distortion in the perspective of the robot.}
%   \label{dist_data}
% \end{figure}

\begin{figure}[t]
  \centering
  \includegraphics[width=8cm]{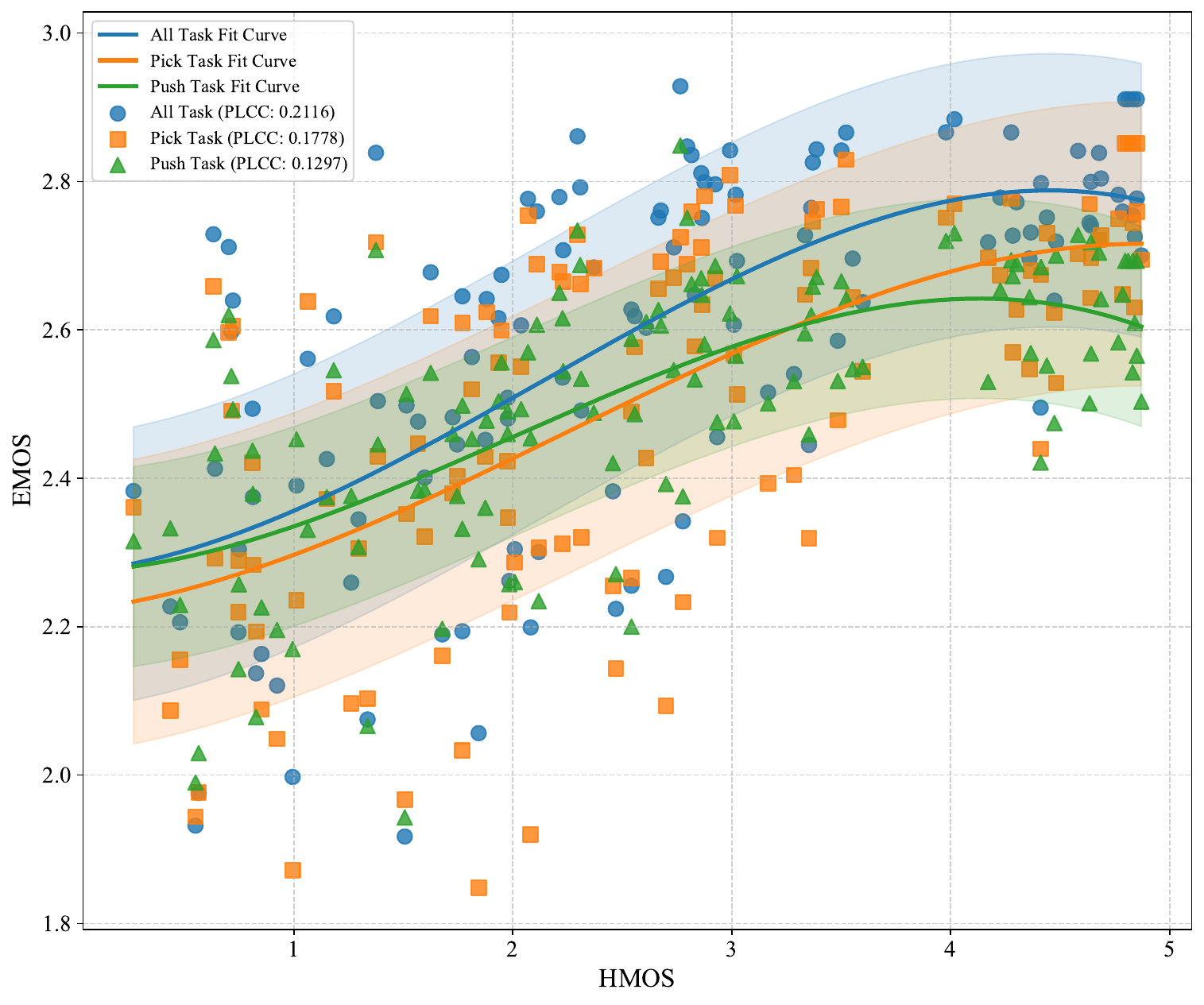}
  \caption{Scatter plots and polynomial fitted curves of embodied MOS (EMOS) and human MOS (HMOS) on the EPD database. The dotted lines represent the 95$\%$ prediction
interval.}
  \label{corelation}
\end{figure}

With the intention of exploring the differences between embodied intelligence and humans with respect to IQA, experiments compared the correlation between human subjective scores and scores on embodied tasks. The correlation is quantified by the Pearson Linear Correlation Coefficient (PLCC). A human subjective experiment is conducted to obtain the MOS of the EPD database. Images obtained from the embodied task were evaluated by 15 experienced experts to obtain a human MOS. Fig. \ref{corelation} illustrates the correlation between humans and robots on the EPD database. The correlation between human and embodied intelligence preference scores for images is low, with a correlation coefficient of 0.2116, 0.1778, 0.1297 respectively. This also confirms that there is a gap between machines and humans regarding image quality, and therefore the traditional HVS-based IQA method not be adapted to embodied intelligence.

\section{Proposed Methed}
The task of image quality assessment for embodied AI face specific challenges in terms of real-time model perception. On one hand, the computational resources of hardware platforms on which embodied robots are limited, requiring IQA models to be lightweight and efficient. On the other hand, the visual systems of embodied AI are more concerned with physical world information that directly impacts their decision and execution. For embodied robots, the object texture details, scene structure, and potential motion blur are more important, rather than the aesthetic quality preferred by the human visual system.

Towards these requirements, this paper proposes a novel multi-scale attention embodied image quality assessment, called MA-EIQA. This network utilizes an efficient multi-scale feature encoder to capture rich features crucial for robotic tasks. It also introduces an embodied attention module to dynamically focus on the most informative feature regions. Finally, a linear layer is used to regress an accurate quality score.

\subsection{Multi-scale Attention Embodied Image Quality Assessment}
\begin{figure*}[t]
  \centering
  \includegraphics[width=18cm]{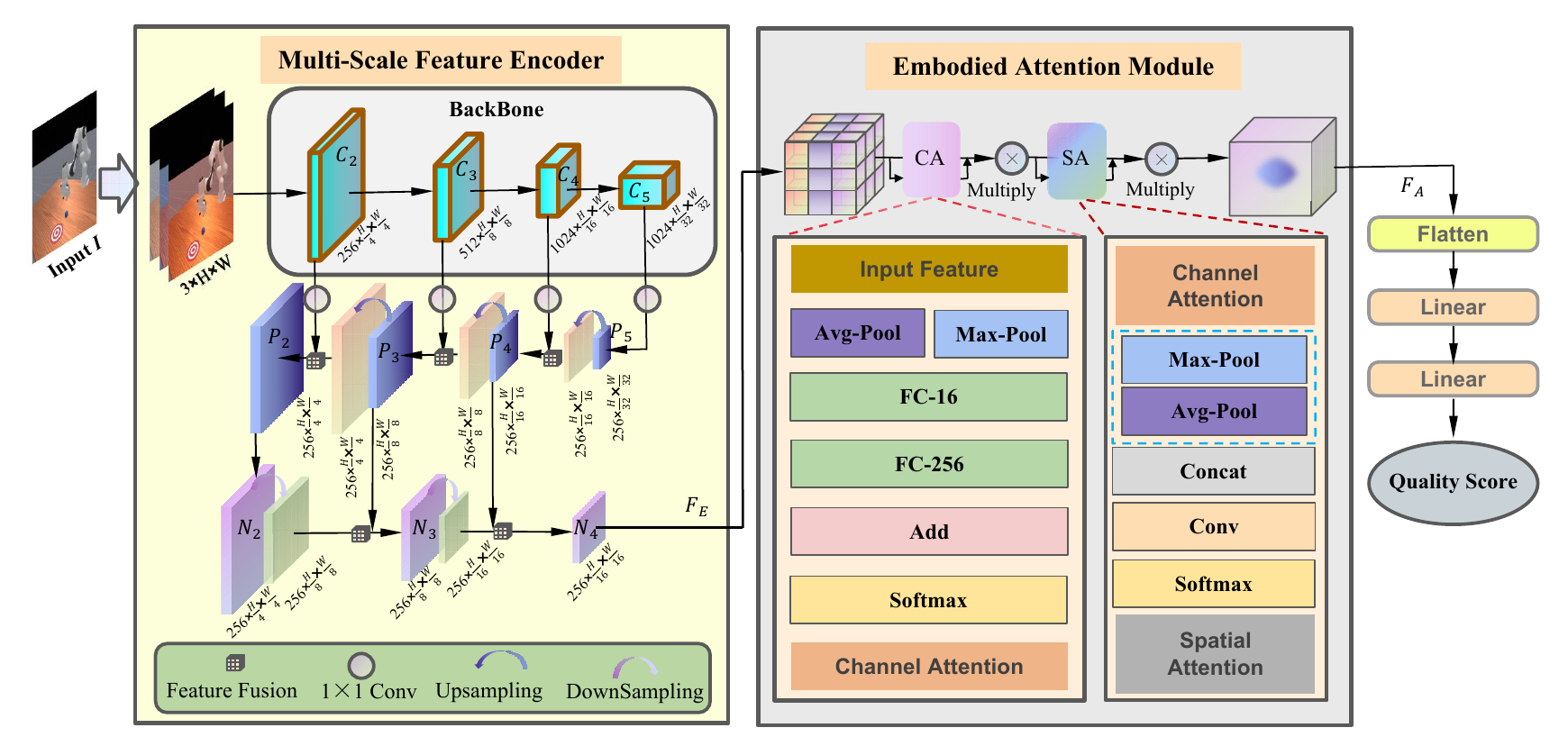}
  \caption{Framework of the proposed method. MA-EIQA is the first embodied No-Reference IQA model, which consists of two components: multi-scale feature extraction and embodied attention module. The input image $I$ goes through feature extraction, and then $F_{E}$ is fed into the attention module. Finally, the extracted features $F_{A}$ are subjected to a linear layer to obtain the final prediction score.}
  \label{network}
\end{figure*}

The overall architecture of the proposed MA-EIQA network is shown in Fig. \ref{network} and consists of two main components: a multi-scale feature encoding module and an embodied attention module.

The specific workflow is as follows:
First, an input image $I$ is fed into a pre-trained ResNet50 \cite{resnet} backbone to extract foundational, multi-level visual features. Subsequently, the feature maps from different stages of the ResNet50 are passed to the multi-scale feature encoder. Inspired by the design of the Path Aggregation Network (PANet) \cite{PANET}, we generates an information-rich, aggregated feature map $F\_{E}$ through bi-directional path fusion.

Next, $F\_{E}$ is sent to the embodied attention module, which refines the features through a series of channel and spatial attention mechanisms to produce an attention-weighted feature map $F\_{A}$.

Finally, the feature map $F\_{A}$ is flattened and passed through two consecutive fully connected linear layers to regress the final image quality score $y$.

The loss function is used to measure the difference between the model predicted score $y$ and the DMOS, denoted as $\hat{y}$, which is annotated by an embodied "expert". Given its penalty for larger errors and its smoothness in gradients, we select the Mean Squared Error (MSE Loss) as the objective function for model optimization. Its definition is as follows:

\begin{equation}
L_{MSE}=\frac{1}{N}\sum_{i=1}^N(y_i-\hat{y}_i)^2 ,
\end{equation}
where $N$ is the number of samples in a batch, $y_i$ is the predicted score for the $i$-th image, and $\hat{y}_i$ is the corresponding ground-truth DMOS.

\subsection{Multi-Scale Feature Encoder}
When interacting with the physical world, embodied AI agents need to understand both the macro-level layout (semantic information) and micro-level details (texture, edge information) of a scene. In embodied manipulation tasks, it is crucial to perceive both the semantic attributes of the task and learn useful information from textural features. A single-scale feature map struggles to satisfy both of these needs simultaneously.

To address this issue, we propose an efficient multi-scale feature encoder. Aiming to achieve deep, cross-scale feature fusion with a lightweight manner. The central idea is to aggregate features from different levels through both a bottom-up and a top-down path. This helps to effectively fuse high-level semantic information with low-level precise localization details. The ResNet50 \cite{resnet} is selected as the backbone because it can provide rich multi-scale features.

The input features are from different stages of the ResNet50, ${C_2, C_3, C_4, C_5}$, where $C_i$ represents the output of the $i$-th stage. The subsequent process involves two information propagation paths:

\textbf{Top-Down Feature Pyramid Path}. This path is responsible for  transmitting high-level semantic abstractions to lower-level layers. The hierarchical feature pyramid construction initiates from the deepest feature map ($C_5$), an initial feature pyramid ${P_2, P_3, P_4, P_5}$ is generated by progressively upsampling and using 1x1 convolutions (for channel matching) to fuse with lower-level features. This process can be expressed as:
\begin{equation}
P_i=f_{up}(P_{i+1})+T_i(C_i),\quad i\in2,3,4 ,
\end{equation}
where $f_{up}(\cdot)$ denotes an upsampling operation (such as bilinear interpolation), and $\mathcal{T}_i(\cdot)$ represents a 1x1 convolutional layer used for adjusting channel dimensions and feature projection.

\textbf{Bottom-Up Feature Enhancement Path}. To compensate for the potential dilution of low-level feature information that can occur during long-range propagation in the top-down path, an additional bottom-up path is constructed. Starting from the lower-level pyramid feature map ($P_2$), this path utilizes the 3×3 convolutions‌ (stride=2) for hierarchical downsampling. This path progressively enhance higher-level feature maps with rich spatial details from lower layers, producing the final feature pyramid ${N_2, N_3, N_4, N_5}$:
\begin{equation}
N_i=f_{down}(N_{i-1})+P_{i},\quad i\in3,4 ,
\end{equation}
where $N_2 = P_2$, and $f_{down}(\cdot)$ represents a 3x3 convolutional layer with a stride of 2, used for downsampling and feature transformation.

The final output feature $F\_E$ from this module is able to contain both rich textural and edge details as well as high-level semantic information, providing a high-quality input for the subsequent attention module.

\subsection{Embodied Attention Module}
Not all features are equally important for the final quality assessment. To enable the model to adaptively focus on these critical features that are useful to robots, we propose an Embodied Attention Module. The structure of this module is based on the Convolutional Block Attention Module (CBAM) \cite{CBAM}. The module takes the feature $F_E \in \mathbb{R}^{C \times H \times W}$ from the previous stage as input and outputs an attention-weighted feature map $F_A$. This module refines and weights features from both channel and spatial dimensions through a series of two sub-modules: Channel Attention and Spatial Attention.

\textbf{Channel Attention Module}. This module aims to model the importance of different feature channels. It aggregates the spatial information of each channel using both global average pooling (AvgPool) and global maximum pooling (MaxPool). Both are then fed into a shared, lightweight multi-layer perceptron (MLP), and the final channel attention weights $M_c \in \mathbb{R}^{C \times 1 \times 1}$ are generated through a Sigmoid activation function.
\begin{equation}
M_c(F_E)=\sigma(\mathrm{MLP(AvgPool}(F_E))+\mathrm{MLP(MaxPool}(F_E))) ,
\end{equation}
where $\sigma$ denotes the Sigmoid function. The resulting channel attention weights are then broadcast and multiplied element-wise with the original feature map to obtain the channel-weighted feature map $F'$:
\begin{equation}
F^{\prime}=M_c(F_E)\otimes F_E
\end{equation}

\textbf{Spatial Attention Module}. This module follows the channel attention and is designed to identify important regions within the feature map. It first applies both average pooling and max pooling along the channel dimension of the channel-weighted feature map, $F'$, to generate two 2D spatial descriptors. These two descriptors are then concatenated and fused via a standard convolutional layer (i.e., a 7x7 convolution). Finally, a 2D spatial attention map, $M_s \in \mathbb{R}^{1 \times H \times W}$, is generated using a Sigmoid activation function.
\begin{equation}
M_s(F^{\prime})=\sigma(f^{7\times7}([\mathrm{AvgPool}*ch(F^{\prime});\mathrm{MaxPool}*ch(F^{\prime})])),
\end{equation}
where $f^{7 \times 7}$ denotes a convolution operation with a $7 \times 7$ kernel, and $[\cdot;\cdot]$ represents a concatenation operation. Similarly, the resulting spatial attention map is broadcasted and multiplied element-wise with $F'$ to obtain the final refined feature map, $F_A$:
\begin{equation}
F_A=M_s(F^{\prime})\otimes F^{\prime}
\end{equation}

By processing features through the embodied attention module in series, the model is able to suppress irrelevant background and noise, while simultaneously enhancing the feature representations most relevant to image quality. This provides a more focused feature representation for the subsequent quality score regression.

\section{Experiment}
This section describes the setup of the experiment, then analyses the environment of the experiment, and finally evaluates the results of the experiment.

\subsection{Experiment and Environment Setups}
Oriented to the 100 episodes of the embodied task, 25 common image distortions with 5 different levels of intensity are set up, and a total of 12,500 image pairs are acquired. The train/val dataset is divided according to an 8:2 ratio. For each pair of images the overall scoring and the score of the three subtasks (i.e. All Task, Push Task and Pick Task) is annotated by embodied robot. Three metrics, SRCC, Kendall Rank Order Correlation Coefficient (KRCC) and PLCC, are used to evaluate the consistency between objective and subjective quality scores. The SRCC and KRCC represent predictive monotonicity and PLCC represents predictive accuracy.

\begin{figure*}[t] %htbp
  \centering
  \includegraphics[width=18cm]{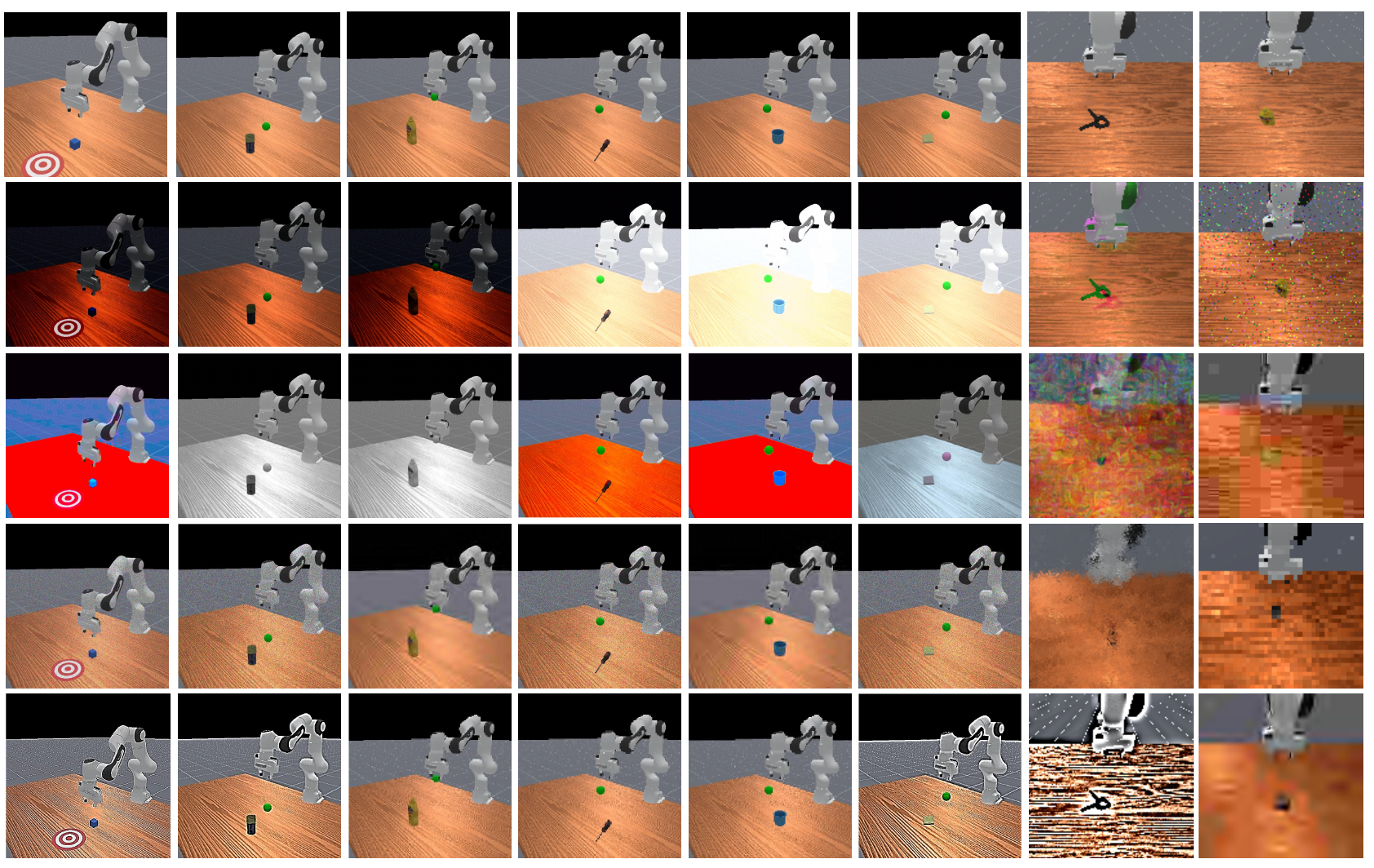}
  \caption{The different scenes that the benchmark can be supported. The EPD support push and pick two major types of tasks. A wide range of different objects and scenes can be flexibly changed. The first row demonstrates the reference undistorted image, while rows 2-5 demonstrate the corresponding distorted images. The experiment rendered various task scenarios from a third-person perspective. The model mainly takes the front view as input, such as 7-8 columns.}
  \label{scenes}
\end{figure*}

For algorithm evaluation, we utilize 16 mainstream IQA methods for the comparison on EPD dataset. These methods have achieved commendable results in terms of IQA methods based on the human visual system. In addition to comparing traditional PSNR and SSIM \cite{ssim}, the full reference IQA methods include AHIQ \cite{AHIQ}, PieAPP \cite{Pieapp}, CKDN \cite{CKDN}, DISTS \cite{DISTS}, LPIPS \cite{LPIPS}, TOPIQFR \cite{TOPIQ}, IQT \cite{IQT}. The no-reference methods include CLIPIQA \cite{CLIPIQA}, DBCNN \cite{DBCNN}, TOPIQNR \cite{TOPIQ}, MANIQA \cite{Maniqa}, TempQT \cite{TempQT}, HyperIQA \cite{HyperIQA}, QualiCLIP \cite{QualiCLIP}. We trained the algorithms on the EPD train dataset based on the IQA-PyTorch \cite{pyiqa} platform, and test in val dataset. We evaluated each algorithm on three evaluation metrics on three subdatasets. The hardware platform on which the experiments is 8$\times$NVIDIA RTX3090 24G GPUs.

It is important to note that the EPD benchmark is different from the traditional paradigm. Instead of offline collection of various types of images, the EPD are oriented towards an embodied scene. A robot needs to constantly perceive its surroundings as it completes its task. We capture images from a monocular camera mounted on the robot as it performs its task. Moreover, the collected images are primarily for the robot and not for humans. The examples of diverse scenes are shown in Fig. \ref{scenes}.
\begin{figure}[t] %htbp
  \centering
  \includegraphics[width=9cm]{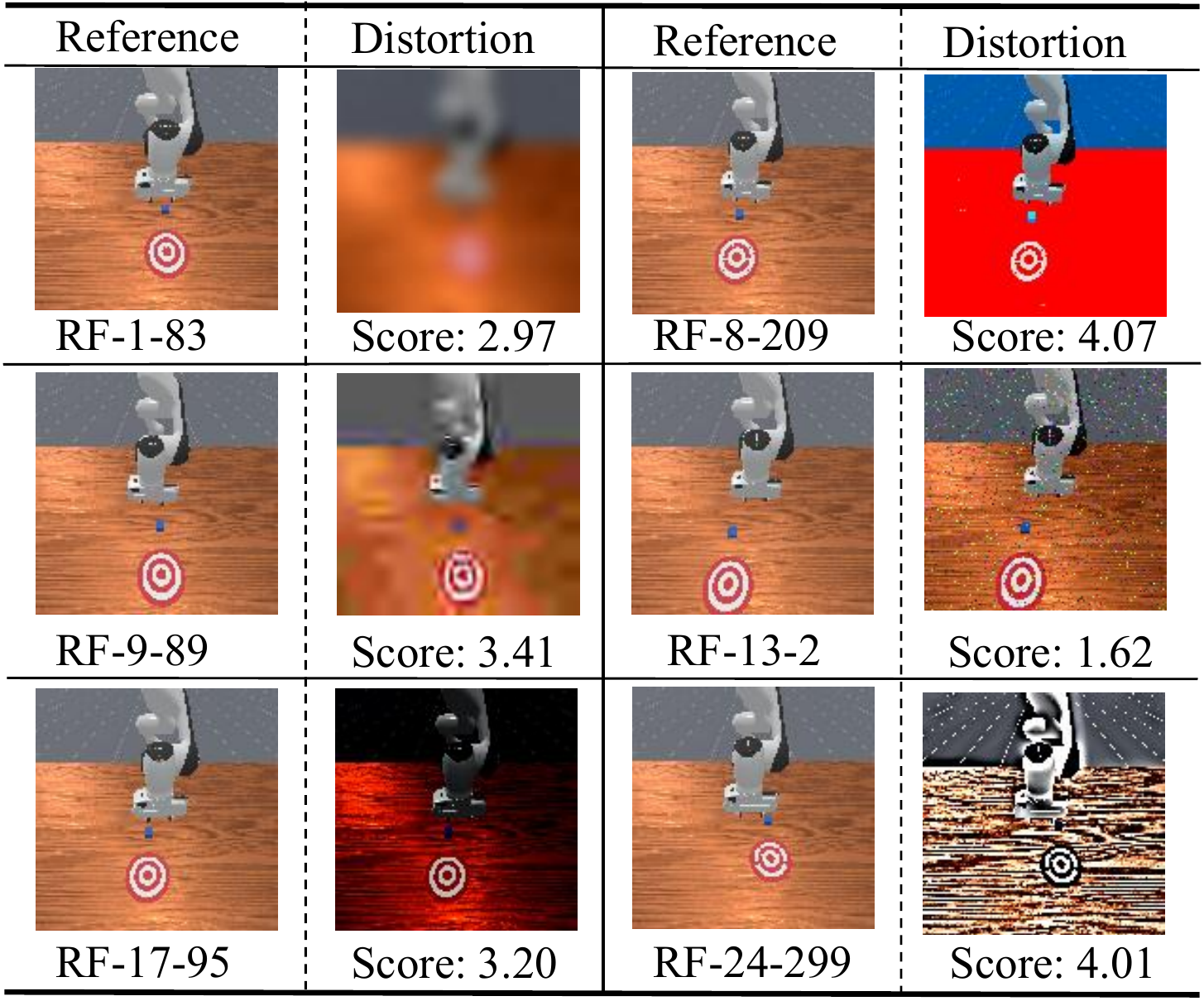}
  \caption{The first and third columns are the reference images, and the second and fourth columns correspond to the distorted images and scores. RF-X-X denotes the file name of the example image in the EPD benchmark.}
  \label{env_exp}
\end{figure}

\begin{table*}[thbp]  %htbp
  \centering  
  \caption{Comparison of 16 IQA methods for BL (Baseline), FR (Full Reference), NR (No reference) respectively on EPD benchmarks. In FR and NR, the Best results are marked in \CLA{Orange} while the second best results are in \CLB{Blue}}  
  \setlength{\tabcolsep}{1.2mm}{
  \begin{tabular}{cl|rrrrrrrrr|rr}  
    \toprule  
    \multirow{2}{*}{Type} & \multirow{2}{*}{Metric} & \multicolumn{3}{c}{All Tasks} & \multicolumn{3}{c}{Push Task} & \multicolumn{3}{c}{Pick Task} & \multirow{2}{*}{Params} & \multirow{2}{*}{Architecture} \\  
    \cmidrule(lr){3-5} \cmidrule(lr){6-8} \cmidrule(lr){9-11}  
    & & SRCC$\uparrow$ & KRCC$\uparrow$ & PLCC$\uparrow$ & SRCC$\uparrow$ & KRCC$\uparrow$ & PLCC$\uparrow$ & SRCC$\uparrow$ & KRCC$\uparrow$ & PLCC$\uparrow$ \\  
    \midrule
    \multirow{2}[2]{*}{BL} & PSNR  & 0.1660 & 0.1107 & 0.1651 & 0.0928 & 0.0656 & 0.0969 & 0.1472 & 0.0978 & 0.1429 & - & Pixel-based \\
          & SSIM (TIP2004) \cite{ssim}  & 0.2347 & 0.1572 & 0.2407 & 0.1392 & 0.0931 & 0.1386 & 0.2085 & 0.1390 & 0.2111 & - & Pixel-based \\
    \midrule
    \multirow{6}[2]{*}{FR} 
          & PieAPP (CVPR2018) \cite{Pieapp} & 0.2538 & 0.1707 & 0.1942 & 0.1863 & 0.1243 & 0.1414 & 0.2143 & 0.1438 & 0.2040 & 68.38 M & CNN-based \\
          & LPIPS (CVPR2018) \cite{LPIPS}  & 0.1886 & 0.1254 & 0.2062 & 0.1110  & 0.0736 & 0.1213 & 0.1915 & 0.1279 & 0.1898 & 14.72 M & CNN-based \\
          & CKDN (ICCV2021) \cite{CKDN}  & 0.3671 & 0.2504 & 0.3435 & 0.2511 & 0.1688 & 0.1836 & 0.4310 & 0.2968 & 0.4118 & 13.38 M & CNN-based \\
          & IQT (CVPR2021) \cite{IQT}   & 0.5123 & 0.3568 & 0.5315 & 0.3654 & 0.2493 & 0.3838 & 0.5460 & 0.3839 & 0.5618 & 57.72 M & Transformer \\
          & AHIQ (CVPR2022) \cite{AHIQ}  & 0.3068 & 0.2080 & 0.3245 & 0.2553 & 0.1722 & 0.2647 & 0.3315 & 0.2254 & 0.3172 & 139.30 M & Transformer \\
          & DISTS (TPAMI2022) \cite{DISTS} & 0.0737 & 0.0495 & 0.0641 & 0.0625 & 0.0415 & 0.0685 & 0.0932 & 0.0627 & 0.0946 & 14.72 M & CNN-based \\
          & TOPIQ-FR (TIP2024) \cite{TOPIQ} & 0.2923 & 0.1974 & 0.3052 & 0.1425 & 0.0952 & 0.1471 & 0.2332 & 0.1560 & 0.2417 & 36.04 M & CNN-based \\
    \midrule
    \multirow{6}[2]{*}{NR} & HyperIQA (CVPR2020) \cite{HyperIQA} & 0.4502 & 0.3081 & 0.4487 & 0.4604 & 0.3155 & 0.4686 & 0.4762 & 0.3324 & 0.4943 & 27.38 M & CNN-based \\
          & DBCNN (TCSVT2021) \cite{DBCNN} & 0.2922 & 0.1971 & 0.2981 & 0.2851 & 0.1922 & 0.2894 & 0.3371 & 0.2295 & 0.3349 & 15.31 M & CNN-based \\
          & MANIQA (CVPR2022) \cite{Maniqa} & \CLB{0.5526} & \CLB{0.3855} & \CLB{0.5716} & 0.4874 & 0.3368 & 0.5032 & \CLB{0.5666} & \CLB{0.4032} & \CLB{0.5956} & 135.75 M & Transformer \\
          & CLIPIQA (AAAI2023) \cite{CLIPIQA} & 0.3471 & 0.2344 & 0.3222 & 0.2890 & 0.1941 & 0.2748 & 0.3322 & 0.2265 & 0.3152 & 102.02 M & Transformer \\
          & TempQT (TMM2024) \cite{TempQT} & 0.5500 & 0.3840 & 0.5620 & \CLA{0.5380} & \CLA{0.3760} & \CLA{0.5460} & 0.5470 & 0.3880 & 0.5780 & 87.45 M & Transformer \\
          & TOPIQ-NR (TIP2024) \cite{TOPIQ} & 0.2139 & 0.1433 & 0.2251 & 0.1081 & 0.0720 & 0.1227 & 0.1959 & 0.1308 & 0.2009 & 45.20 M & CNN-based \\
          & QualiCLIP (CVPR2025) \cite{QualiCLIP} & 0.3728 & 0.2528 & 0.3669 & 0.3343 & 0.2261 & 0.3213 & 0.3669 & 0.2506 & 0.3713 & 102.12 M & Transformer \\
    \midrule
    \rowcolor{lightgray}
          & MA-EIQA (our) & \CLA{0.5755} & \CLA{0.4032} & \CLA{0.5836} & \CLB{0.4933} & \CLB{0.3408} & \CLB{0.5040} & \CLA{0.5738} & \CLA{0.4084} & \CLA{0.5971} & 48.83 M & CNN-based \\
    \bottomrule  
  \end{tabular}}  
  \label{tab}  
\end{table*} 

For the experiments, the ManiSkill benchmark \cite{ManiSkill3} is chosen to set up the experimental environment, which is a development platform for reinforcement learning tasks and imitation learning with embodied intelligence based on the SAPIEN simulator \cite{SAPIEN}. The experiment is set up with two subtasks of pushing and picking a box, and the goal is to complete the specified tasks through reinforcement learning, i.e., push the box to the center of the sign and pick up the box. Each episodes represents one complete interaction of the robotic arm with the environment in different scenarios, and the experiment set 50 steps of actions for each episodes. Incidentally, reward values are given for each step.

To explore the capability of the robot to task completion under different distorted images, the type and intensity of distortion set in an episodes is consistent. The final task completion performance is measured by the average reward value. Furthermore, it can score the images captured by the robot while completing the embodied task. The example scores of several distorted images are shown in Fig. \ref{env_exp}. Limited by the size and computational resources of the available reinforcement learning models, the image inputs captured by the robot are all 128$\times$128 pixels.

\subsection{Experiment Results on EPD}

\begin{table*}[h]  %htbp
  \centering  
  \caption{Score distribution for 25 common image distortions at level 5 distortion for embodied tasks. MEAN represents the average score for embodied tasks, and std represents the standard deviation of stability. The Best results are marked in \CLA{Orange} while the worst results are in \CLB{Blue}.}  
  \setlength{\tabcolsep}{1.0mm}{
  \begin{tabular}{c|l|cccccccccc|cc}  
    \toprule  
    \multirow{2}{*}{Distortion Type} & \multirow{2}{*}{Distortion} & \multicolumn{2}{c}{Level 1} & \multicolumn{2}{c}{Level 2} & \multicolumn{2}{c}{Level 3} & \multicolumn{2}{c}{Level 4} & \multicolumn{2}{c}{Level 5} & \multicolumn{2}{c}{Average} \\  
    \cmidrule(lr){3-4} \cmidrule(lr){5-6} \cmidrule(lr){7-8} \cmidrule(lr){9-10} \cmidrule(lr){11-12} \cmidrule(lr){13-14} 
    & & Mean & Std & Mean & Std & Mean & Std & Mean & Std & Mean & Std & Mean & Std \\  
    \midrule
    \multirow{3}{*}{Blurs (D.1)} & Gaussian blur  & \CLA{2.8183} & \CLA{0.5226} & 2.7588 & 0.8119 & 2.6850 & 0.8089 & 2.5604 & 0.7954 & 2.2156 & 0.8319 & 2.6076 & 0.7541 \\
          & Lens blur  & 2.7305 & 0.8097 & 2.7021 & 0.7760 & 2.4597 & 0.7622 & 2.2835 & 0.7404 & 2.1969 & 0.8697 & 2.4745 & 0.7916 \\
          & Motion blur  & \CLA{2.8183} & \CLA{0.5226} & 2.7788 & 0.7892 & 2.7000 & 0.7819 & 2.7547 & 0.7417 & 2.5791 & 0.8173 & \CLA{2.7262} & \CLA{0.7305} \\
    \midrule
    \multirow{5}{*}{Color distortions (D.2)} 
          & Color diffusion  & 2.5685 & 0.8049 & \CLB{1.9424} & 0.8155 & \CLB{2.0131} & 0.7804 & \CLB{1.9942} & 0.7382 & \CLB{1.9553} & 0.7599 & \CLB{2.0947} & 0.7798 \\
          & Color shift  & 2.6290 & 0.7970 & 2.4354 & 0.7373 & 2.4082 & 0.7590 & 2.4405 & 0.7676 & 2.4545 & 0.7972 & 2.4735 & 0.7716 \\
          & Color quantization  & 2.7545 & 0.7629 & 2.7202 & 0.7623 & 2.6684 & 0.7791 & 2.6959 & 0.7908 & 2.4486 & 0.7456 & 2.6575 & 0.7681 \\
          & HSV saturation  & 2.3168 & \CLB{0.8420} & 2.2403 & 0.8590 & 2.2804 & 0.7777 & 2.2837 & 0.8094 & 2.1825 & 0.8543 & 2.2608 & \CLB{0.8285} \\
          & Lab saturation  & \CLA{2.8183} & \CLA{0.5226} & 2.2509 & 0.8297 & 2.1908 & 0.7793 & 2.1862 & 0.7977 & 2.2440 & 0.8146 & 2.3380 & 0.7488 \\
    \midrule
    \multirow{2}{*}{Compression (D.3)} 
          & JPEG2000 compression  & 2.7520 & 0.7035 & 2.6185 & 0.7749 & 2.6097 & 0.7316 & 2.5100 & 0.7677 & 2.1363 & 0.8032 & 2.5253 & 0.7562 \\
          & JPEG compression  & 2.6127 & 0.8168 & 2.5582 & 0.7760 & 2.5606 & 0.7970 & 2.4506 & 0.7461 & 2.3529 & 0.8164 & 2.5070 & 0.7905 \\
    \midrule
    \multirow{5}{*}{Noise (D.4)} 
          & White noise  & 2.7048 & 0.8208 & 2.6860 & 0.7839 & 2.7622 & 0.7976 & 2.7136 & 0.7424 & 2.5121 & 0.8359 & 2.6758 & 0.7961 \\
          & Color noise  & 2.7789 & 0.7491 & \CLA{2.8338} & 0.7379 & 2.7198 & 0.7771 & 2.6389 & 0.8048 & 2.5842 & \CLA{0.7263} & 2.7111 & 0.7590 \\
          & Impulse noise  & 2.7791 & 0.7460 & 2.7432 & \CLA{0.7363} & 2.6892 & 0.7864 & 2.4918 & 0.8310 & 2.5504 & 0.7533 & 2.6507 & 0.7706 \\
          & Multiplicative noise  & 2.6600 & 0.8230 & 2.7945 & 0.7941 & 2.7574 & 0.8312 & 2.6567 & 0.7343 & 2.5809 & 0.7816 & 2.6899 & 0.7928 \\
          & Gaussian Denoise & 2.6776 & 0.7739 & 2.5497 & 0.8320 & 2.4636 & 0.8194 & 2.3594 & 0.7262 & 2.3794 & 0.7772 & 2.4860 & 0.7857 \\
    \midrule
    \multirow{3}{*}{Brightness change (D.5)} & Brighten  & 2.6550 & 0.8400 & 2.6327 & \CLB{0.8671} & 2.6563 & 0.7681 & 2.4700 & 0.8050 & 2.3525 & 0.8251 & 2.5533 & 0.8211 \\
          & Darken & 2.7048 & 0.7788 & 2.6715 & 0.7927 & 2.5773 & 0.8330 & 2.4169 & 0.8329 & 2.0651 & 0.8088 & 2.4871 & 0.8092 \\
          & Mean shift & 2.6778 & 0.7988 & 2.6805 & 0.8163 & \CLA{2.8183} & \CLA{0.5226} & 2.7004 & 0.7926 & 2.4519 & 0.8061 & 2.6658 & 0.7473 \\
    \midrule
    \multirow{5}{*}{Spatial distortions (D.6)} 
          & Jitter & 2.7242 & 0.7540 & 2.7016 & 0.7544 & 2.6773 & 0.8158 & 2.6427 & 0.7981 & \CLA{2.6579} & \CLB{0.8887} & 2.6807 & 0.8022 \\
          & Non-eccentricity patch & 2.4136 & 0.7920 & 2.3908 & 0.7632 & 2.3191 & 0.8621 & 2.1849 & \CLB{0.8598} & 2.3457 & 0.7911 & 2.3308 & 0.8136 \\
          & Pixelate & 2.7576 & 0.7152 & 2.7173 & 0.8597 & 2.6423 & 0.7636 & \CLA{2.7744} & 0.8003 & 2.5421 & 0.7481 & 2.6867 & 0.7774 \\
          & Quantization & 2.7188 & 0.8034 & 2.7570 & 0.8052 & 2.6291 & 0.8201 & 2.6491 & 0.7754 & 2.5790 & 0.7973 & 2.6666 & 0.8003 \\
          & Color block & \CLB{2.2128} & 0.7638 & 2.2460 & 0.8536 & 2.0815 & 0.7723 & 2.1217 & 0.8157 & 2.1592 & 0.8185 & 2.1643 & 0.8048 \\
    \midrule
    \multirow{2}{*}{Sharpness and contrast (D.7)} 
          & High sharpen & 2.6261 & 0.8346 & 2.5468 & 0.8290 & 2.4876 & 0.7981 & 2.4390 & 0.8280 & 2.3688 & 0.8509 & 2.4937 & 0.8281\\
          & Contrast change & 2.5860 & 0.7816 & 2.5314 & 0.7457 & 2.6483 & \CLB{0.8699} & 2.6848 & \CLA{0.6908} & 2.4076 & 0.8465 & 2.5716 & 0.7869 \\
    \bottomrule  
  \end{tabular}}  
  \label{distortion}  
\end{table*} 

Table \ref{tab} demonstrates the performance of IQA methods on the EPD. As can be seen from the table, the traditional IQA methods based on the human visual system do not perform well in the IQA of the robot visual system. It also proves that the perspective of embodied robots is not consistent with the image assessment methods in human perspective. For image quality assessment from the robotic viewpoint the algorithms cannot be designed simply with the human visual system.

Traditional FR-based methods do not perform well because they differ from human assessment of images, whereas NR methods can extract more useful features from a robotic perspective.

\begin{table}[t]
  \centering
  \caption{Comparison of DMOS in different image distortion for 6 EAI participant, which normalised to between 0 and 5. The D.1 to D.7 denote seven different types of image distortions, namely blur, colour distortions, compression, noise, brightness change, spatial distortions, sharpness and contrast. \textbf{Bold} data indicate the average best result}
  \setlength{\tabcolsep}{1.75mm}{
  \begin{tabular}{c|cccccc|c}
  \toprule
  Dist & EAI\_1 & EAI\_2 & EAI\_3 & EAI\_4 & EAI\_5 & EAI\_6 & Mean \\ 
  \midrule
  D.1 & 2.5691 & \textbf{2.7287} & 2.5045 & 2.3813 & 2.5673 & \textbf{2.6754} & 2.5714 \\
  D.2 & 2.3490 & 2.2678 & 2.4624 & 2.3702 & 2.3559 & 2.2895 & 2.3491 \\
  D.3 & 2.4403 & 2.6442 & 2.4530 & \textbf{2.6420} & 2.4266 & 2.3353 & 2.4902 \\
  D.4 & \textbf{2.6412} & 2.6861 & 2.5036 & 2.5576 & \textbf{2.6234} & 2.6377 & \textbf{2.6083} \\
  D.5 & 2.5941 & 2.5269 & 2.5403 & 2.6025 & 2.4416 & 2.5302 & 2.5327 \\
  D.6 & 2.4216 & 2.4247 & 2.4979 & 2.5178 & 2.5305 & 2.4909 & 2.4806 \\
  D.7 & 2.5360 & 2.2911 & \textbf{2.5701} & 2.5332 & 2.5359 & 2.5674 & 2.5056 \\ 
  \bottomrule 
  \end{tabular}}
  \label{eai-mos}
  \end{table}

Current state-of-the-art methods can only achieve performance below 0.6. In the FR method, IQT \cite{IQT} is able to achieve SRCC and PLCC above 0.5 on All Task dataset. In the NR method, MANIQA \cite{Maniqa} and TempQT \cite{TempQT} are able to achieve SRCC and PLCC above 0.5. Models based on the Transformer architecture have higher average performance. Overall, the NR methods perform slightly better than the FR methods. Existing methods are essentially model the human visual system, while less attention is paid to the robotic visual system. The NR approaches can extract more useful features from a robotic perspective. This phenomenon further reveals a fundamental characteristic of embodied AI systems: $\textbf{only} \  \textbf{the} \  \textbf{machine} \  \textbf{that} \  \textbf{understands} \  \textbf{the} \  \textbf{machine} \  \textbf{better}$.

In addition, the performance of each method varies in different subtasks. The Pick task has a higher performance than the Push task, while the All task has the highest performance. Embodied IQA still has significant optimization space.

Table \ref{distortion} shows the scores of 25 common image distortions in embodied tasks of level 5 normalized to (0, 5). The mean values in the table reflect the overall task performance, while the standard deviation std reflects the impact of image distortion on task robustness. To validate the influence of task types and participants on task results, six embodied intelligent robot subjects are selected, experiments are conducted as outlined in Table \ref{eai-mos}. Just as different human participants have varying evaluations of images, so too do different embodied robots. Each EAI assesses image quality differently in different image distortion environments. Based on the results in the table, where findings are summarized as follows:
\begin{itemize}
\item Different image distortions have varying degrees of impact on embodied tasks. Motion blur causes minimal disruption to embodied tasks, while color distortion causes maximum disruption.
\item Image distortion with minimal motion blur affects the robustness of embodied tasks, while color distortion reduces robustness.
\item The noise category of image distortion has the least impact on embodied tasks, while color distortion has the most severe impact on embodied tasks.
\item Embodied tasks have different sensitivities to changes in image distortion strength. JPEG2000 compression and Color diffusion are the two types of image distortion that are most sensitive to embodied tasks, while multiplicative noise is the least sensitive.
\end{itemize}
% There are two participants each who scored high on noise and spatial distortion. One participant each scored the highest on compression, sharpness and contrast distortion. This also shows that there is variability in the image quality assessment by different EAI individuals.

\subsection{Performance analysis of MA-EIQA}
Table \ref{tab} demonstrates the performance of the proposed MA-EIQA. It outperforms all FR methods in the table and achieves state-of-the-art performance among NR methods. MA-EIQA is a lightweight network based on CNN, with performance that can be close to and even outperform existing models based on the Transformer architecture. On the All Task and Pick Task datasets, the means of the three metrics SRCC, KRCC, and PLCC are 2.24$\%$, 0.86$\%$, and 2.39$\%$ higher than those of MANIQA, respectively. The multi-scale feature extraction and embodied attention modules enable the model to learn more features that are useful for embodied tasks.

Different from offline network inference, embodied robots in industrial settings have limited computing resources. In the practice of deep learning deployment on edge-side embedded devices, CNN architectures benefit from long-term hardware adaptation and currently dominate traditional edge computing platforms. Transformer architectures, on the other hand, rely on matrix multiplication acceleration units in next-generation NPUs, and their industrial support varies. Image quality assessment should not consume excessive computational resources throughout the embodied task pipeline. Consequently, the adoption of a lightweight network architecture becomes imperative. Table. \ref{tab} compares the model parameters of our proposed method with those of the mainstream IQA methods. MA-EIQA is based on a CNN architecture with 48.83 million parameters. The number of parameters is reduced by 64.03 $\%$ compared to the Transformer-based MANIQA \cite{Maniqa} and by 44.16 $\%$ compared to TempQT \cite{TempQT}.

\subsection{Ablation Study}

To systematically evaluate the effectiveness of each core component of the proposed MA-EIQA, we perform the ablation studies on EPD dataset. This experiment aims to verify the specific contributions of the multi-scale feature extraction module (MS) and the embodied attention module (EA) to the performance of model. All model variants used identical training strategies, hyperparameter configurations, and data splits to ensure a fair comparison.
There the following four model for comparison:

\textbf{Baseline}: The backbone without the modules—the multi-scale feature encoder (MS) and the embodied attention module (EA). 

\textbf{Baseline+MS}: The backbone with the modules—the multi-scale feature encoder (MS) but without the embodied attention module (EA).

\textbf{Baseline+EA}: The backbone with the embodied attention module (EA) but without the modules—the multi-scale feature encoder (MS).

\textbf{MA-EIQA}: Our comprehensive approach includes the modules—the multi-scale feature encoder (MS) and the embodied attention module (EA).

The four model are tested 10 times and the mean results are shown in Table. \ref{AB}. 

By comparing the results of the Baseline and Baseline+MS, which demonstrates the necessity of multi-scale feature encoder in IQA tasks and the effectiveness of the embodied attention module. SRCC, KRCC, and PLCC increased by 6.55$\%$, 7.28$\%$, and 6.61$\%$, respectively. The reason for this is that the baseline model only uses the high-level features from the final stage of ResNet50. While these features contain rich global semantic information, their low resolution results in the loss of a large amount of low-level spatial and textural details crucial for embodied perception during the multiple downsampling steps. The multi-scale feature encoding module effectively fuses high-level abstract semantics with fine-grained structural features. This allows the model to not only understand the overall content of the image but also to accurately capture the key local distortions that determine image quality. 

Furthermore, the embodied attention module can also enhance overall accuracy. SRCC, KRCC, and PLCC increased by 3.85$\%$, 3.92$\%$, and 3.45$\%$, respectively. The design of this module helps the robot learn which features are beneficial for the evaluation task. It is important to note that these attention regions are not entirely the same as those of human vision. The combination of MS and EA modules (i.e. MA-EIQA) achieves the best performance. SRCC, KRCC and PLCC increased by 12.03$\%$, 13.17$\%$, and 12.14$\%$ respectively compared to the baseline.

% \begin{figure}[t] %htbp
%   \centering
%   \includegraphics[width=9cm]{legend-epd.pdf}
%   \caption{Radar charts of ablation experiments across various datasets and individual metrics.}
%   \label{radar}
% \end{figure}

\begin{table*}[thbp]  %htbp
  \centering  
  \caption{Comparison of 14 IQA methods for BL (Baseline), FR (Full Reference), NR (No reference) respectively on EPD benchmarks. \textbf{Blod} represent the best results in BL. In FR and NR, the Best results are marked in \CLA{Orange} while the second best results are in \CLB{Blue}}  
  \setlength{\tabcolsep}{2.2mm}{
  \begin{tabular}{l|rrrrrrrrr|rrr}  
    \toprule  
    \multirow{2}{*}{Metric} & \multicolumn{3}{c}{All Tasks} & \multicolumn{3}{c}{Push Task} & \multicolumn{3}{c}{Pick Task} & \multicolumn{3}{c}{Average} \\  
    \cmidrule(lr){2-4} \cmidrule(lr){5-7} \cmidrule(lr){8-10} \cmidrule(lr){11-13}
    & SRCC$\uparrow$ & KRCC$\uparrow$ & PLCC$\uparrow$ & SRCC$\uparrow$ & KRCC$\uparrow$ & PLCC$\uparrow$ & SRCC$\uparrow$ & KRCC$\uparrow$ & PLCC$\uparrow$ & SRCC$\uparrow$ & KRCC$\uparrow$ & PLCC$\uparrow$\\  
    \midrule
     Baseline  & 0.5138 & 0.3557 & 0.5246 & 0.4221 & 0.2892 & 0.4315 & 0.5303 & 0.3733 & 0.5464 & 0.4887 & 0.3394 & 0.5008 \\
          Baseline+MS  & 0.5370 & 0.3742 & 0.5521 & 0.4839 & 0.3338 & 0.4917 & 0.5412 & 0.3842 & 0.5580 & 0.5207 & 0.3641 & 0.5339 \\
          Baseline+EA & 0.5281 & 0.3671 & 0.5347 & 0.4509 & 0.3091 & 0.4612 & 0.5436 & 0.3819 & 0.5585 & 0.5075 & 0.3527 & 0.5181\\
          MA-EIQA  & \textbf{0.5755} & \textbf{0.4032} & \textbf{0.5836} & \textbf{0.4933} & \textbf{0.3408} & \textbf{0.5040} & \textbf{0.5738} & \textbf{0.4084} & \textbf{0.5971} & \textbf{0.5475} & \textbf{0.3841} & \textbf{0.5616} \\
    \bottomrule  
  \end{tabular}}  
  \label{AB}  
\end{table*} 

\begin{figure*}[t] %htbp
  \centering
  \includegraphics[width=18cm]{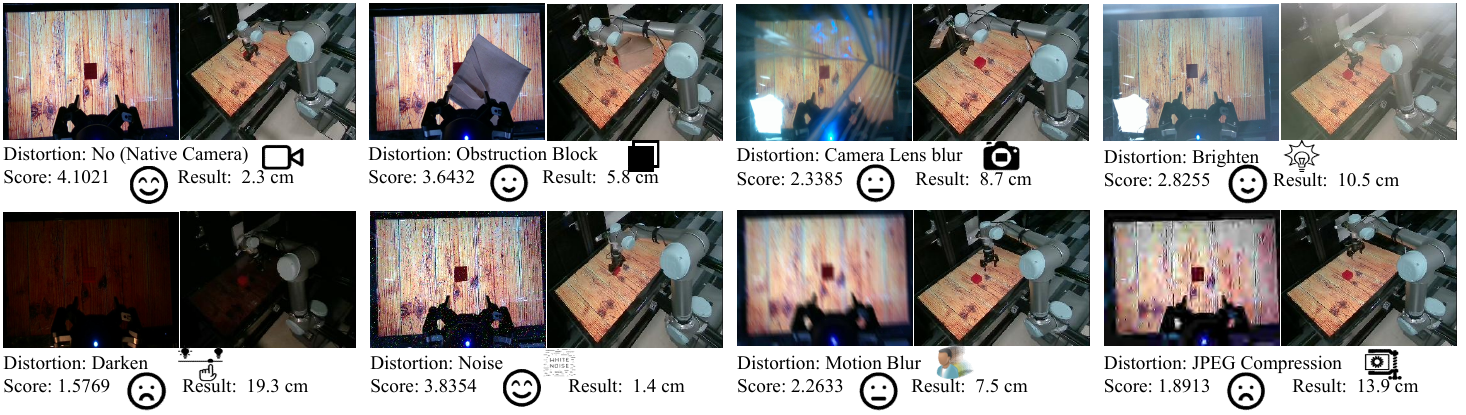}
  \caption{Cases of real-world experiment. The input for the robot is first-person view images with various types of distortion. The third-person perspective shows the results of the execution. We measure the error of task execution by calculating the Euclidean distance from the ground truth of human expert. The score is based on our MA-EIQA prediction score.}
  \label{realdata}
\end{figure*}

\subsection{Real-world Experiment}
We conducted experiments on image distortion and embodied robot operation tasks in real-world scenes. We constructed a flexible experimental scene, including a display platform with arbitrary desktop background changes and variable environmental conditions. In the experiment, the UR5 robot is used as an embodied agent, and the coordinates of the end effector are controlled based on the ROS system. In the experiment, actual working conditions are simulated by changing the external environment and image distortion. Typical scenarios that affect embodied tasks include changes in illumination, object occlusion, lens coverage, and distortion from the camera sensor.

Fig. \ref{realdata} demonstrates the first-person view image input under image distortion conditions and the results after the robotic arm performed the pick task. We use MA-EIQA to perform preference scores on input images. The ability to perform the task is evaluated based on the Euclidean distance between the results executed by the embodied agent and the position where the operation was successfully completed by an experienced human expert. As can be seen from the figure, compression and darkening have a greater impact on embodied tasks. Noise distortion and blocking have minor effects on embodied agents, whereas they severely affect human aesthetics. This also illustrates the difference between humans and robots in terms of image quality perception.

\section{Conslusion}
This paper extends the IQA approach to embodied AI for the first time, and constructs the first robot-oriented image quality assessment database EPD. Experiments demonstrate that there is a gap between the robot vision system and the human vision system, and the current image quality assessment from the human perspective is limited. The application of embodied AI is strongly dependent on the input of visual images, and even far exceeds the human requirements for image quality. In future work, we intend to continue to expand the scope of the embodied preference database to lay the foundation for the development of embodied AI. 

% \section*{Acknowledgments}
% This should be a simple paragraph before the References to thank those individuals and institutions who have supported your work on this article.


% \section{References Section}
% You can use a bibliography generated by BibTeX as a .bbl file.
%  BibTeX documentation can be easily obtained at:
%  http://mirror.ctan.org/biblio/bibtex/contrib/doc/
%  The IEEEtran BibTeX style support page is:
%  http://www.michaelshell.org/tex/ieeetran/bibtex/
 
% You can manually copy in the resultant .bbl file and set second argument of $\backslash${\tt{begin}} to the number of references
%  (used to reserve space for the reference number labels box).

\bibliographystyle{IEEEtran}
\bibliography{bare_jrnl_new_sample4}

% \begin{IEEEbiography}[{\includegraphics[width=1in,height=1.25in,clip,keepaspectratio]{zhangjianbo.JPG}}]{Jianbo Zhang} received the Ph.D. degree in automation from Xinjiang University, Urumqi, China, in 2024, where He is currently pursuing postdoctoral studies at the School of Electronic Information and Electrical Engineering at Shanghai Jiao Tong University, Shanghai, China. 

% His current research interests include embodied artificial intelligence and image quality assessment.
% \end{IEEEbiography}

% \begin{IEEEbiography}[{\includegraphics[width=1in,height=1.25in,clip,keepaspectratio]{YuanLiang-1.png}}]{Liang Yuan} received the Ph.D. degrees in electrical and computer engineering from Ohio State University, Columbus, in 2011. 
    
% Dr. Yuan is currently as the professor of USC-SJTU Institute of Cultural and Creative Industry, Shanghai Jiao Tong University, Shanghai. He was honoured with the National Recruitment Program of Global Experts, in 2014. His research interests include computer vision and imaging processes, robotics, and automation in life science.
% \end{IEEEbiography}

% \begin{IEEEbiography}[{\includegraphics[width=1in,height=1.25in,clip,keepaspectratio]{YuanLiang-1.png}}]{Guoquan Zheng} received the Master's degree in mechanical engineering from Xinjiang University, Urumqi, China. He is currently pursuing the Ph.D. degree with the School of Information Science and Technology, Beijing University of Chemical Technology, Beijing, China.
% His current research interests include 3D human body reconstruction and quality assessment.
% \end{IEEEbiography}

\vfill

\end{document}